%% file: main.tex
\documentclass[10pt,twocolumn,letterpaper]{article}

\usepackage{wacv}              


\definecolor{wacvblue}{rgb}{0.21,0.49,0.74}
\usepackage[pagebackref,breaklinks,colorlinks,allcolors=wacvblue]{hyperref}
\usepackage{pifont}
\newcommand{\cmark}{\ding{51}}%
\newcommand{\xmark}{\ding{55}}%

\title{Does This Moment Justify the Recommendation?\\
Counterfactual Behavior-Grounded Evidence Retrieval\\
for Personalized Video Recommendation}

\author{Xin Liu\\
{\tt\small lxinhit@gmail.com}
}

\begin{document}
\maketitle
\input{sec/abstract}
\input{sec/introduction}
\input{sec/related_work}
\input{sec/task_dataset}

\input{sec/method}
\input{sec/experiments}

\input{sec/limitations}
\input{sec/conclusion}
{\small
\bibliographystyle{ieeenat_fullname}
\bibliography{main}
}

\end{document}

%% file: sec/abstract.tex
\begin{abstract}
Personalized video recommendation predicts user preference at the video level, while temporal video grounding localizes query-relevant moments. However, strong localization does not establish whether the retrieved moment constitutes valid evidence for recommending the video to a particular user. We study \emph{counterfactual behavior-grounded evidence retrieval}, which separates \emph{where} personalized evidence occurs from \emph{whether} such evidence exists and evaluates whether model predictions respond consistently when that evidence is replaced. We introduce \emph{CBGER-10K}, containing 5,000 controlled factual--counterfactual pairs for 3,026 users, where each pair replaces only the focal behavior-supported segment while preserving the user, temporal position, and hard distractors. We further propose \emph{CBGER}, a compact framework that decouples segment-level localization from video-level evidence estimation and learns both through structured counterfactual supervision. CBGER achieves $0.4432$ MRR, $0.6977$ Pair Accuracy, and $0.6987$ Intervention Consistency across five adapted personalized-highlight and temporal-grounding baselines. Notably, compared with QD-DETR, its MRR improvement is not statistically significant, while Pair Accuracy improves by $11.03$ points. These results show that accurate temporal localization does not necessarily imply reliable personalized evidence existence, motivating explicit evaluation of \emph{Whether} alongside \emph{Where}.
\end{abstract}

%% file: sec/introduction.tex
\section{Introduction}
\label{sec:intro}

Video recommendation systems learn from implicit behavioral feedback to rank candidate videos~\cite{ni2023microlens}, but a video-level relevance score provides little temporal evidence about which event supports a recommendation. Temporal video grounding addresses the complementary problem of localizing moments relevant to a language query~\cite{gao2017tall,hendricks2017localizing,lei2021detecting,moon2023query,lin2023univtg}. Standard grounding benchmarks, however, primarily evaluate \emph{where} a query-relevant moment occurs once a query--video relation is defined. Personalized recommendation introduces an additional question: for a particular user's behavior history, does the candidate contain meaningful supporting evidence at all?

\begin{figure*}[t]
  \centering
  \includegraphics[width=\textwidth]{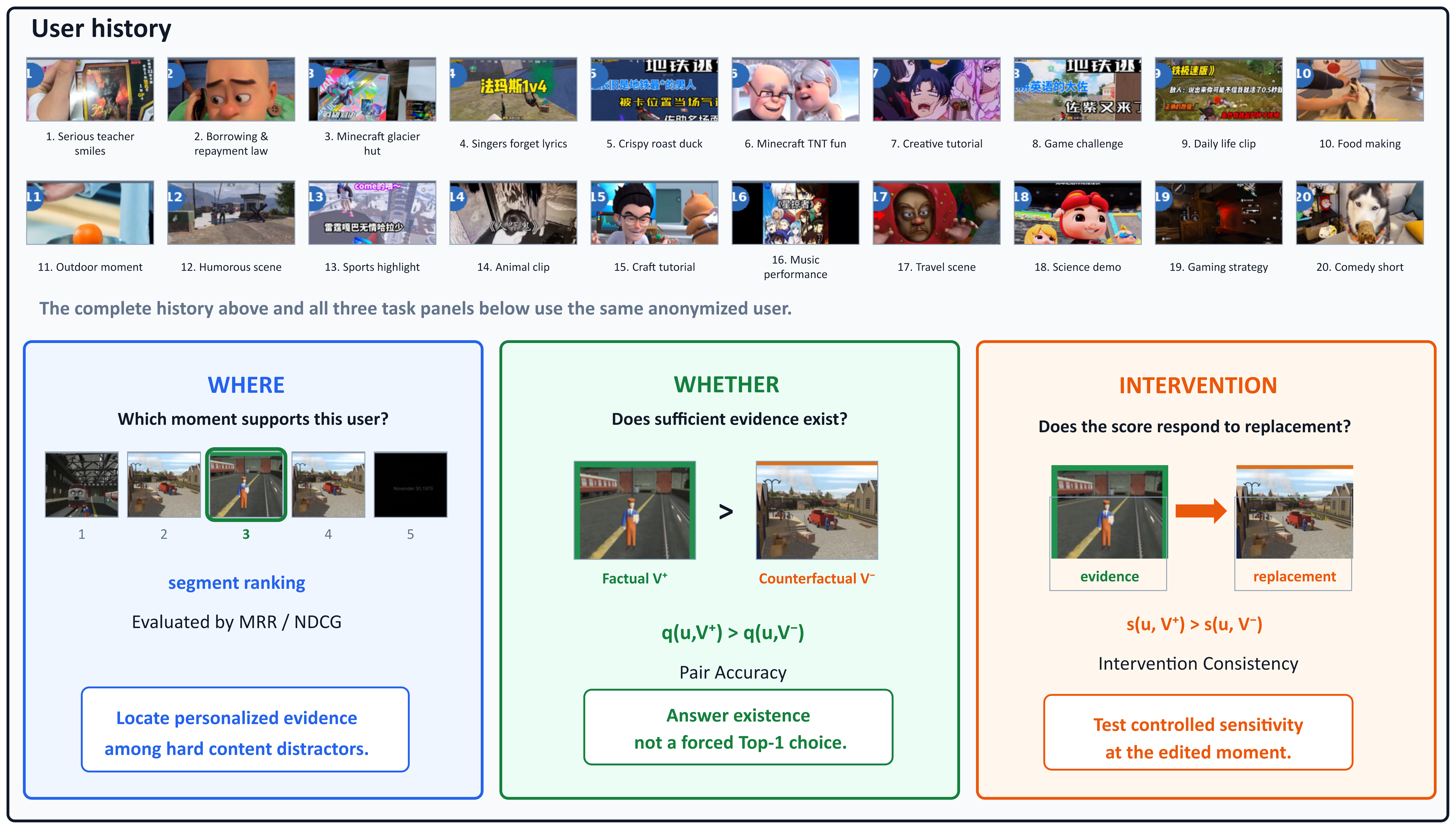}
  \caption{\textbf{CBGER task overview.} A chronological user history defines personalized context. The factual timeline $V^+$ contains a behavior-supported focal segment $e$, while the matched counterfactual $V^-$ replaces only that segment with a behavior-weaker alternative $\tilde e$ at the same temporal position and preserves the surrounding hard distractors. The pair supports evaluation of \textbf{Where}, \textbf{Whether}, and intervention consistency.}
  \label{fig:dataset-overview}
\end{figure*}

This distinction exposes a failure mode that localization metrics alone cannot reveal. A grounding model may rank one segment highest even when it is only the best among weakly relevant alternatives, while a recommendation model may rely on coarse topic compatibility without identifying the event that supports its prediction. We therefore study personalized video evidence along two prediction dimensions and one counterfactual consistency criterion:
\begin{align}
\textbf{Where:}&\quad \text{localize behavior-supported evidence},\nonumber\\
\textbf{Whether:}&\quad \text{whether sufficient evidence is present},\nonumber\\
\textbf{Intervention:}&\quad \text{respond to evidence replacement}.
\end{align}
We operationalize these questions using matched factual--counterfactual videos. As shown in Figure~\ref{fig:dataset-overview}, for a fixed user, a factual timeline contains a behavior-supported focal segment, while its counterfactual replaces only that segment with a semantically related but behavior-weaker alternative at the same temporal position; the remaining hard distractors are unchanged. A faithful model should localize the factual evidence, assign stronger video-level evidence to the factual timeline, and reduce the local score at the edited slot after replacement. This construction tests model dependence on designated evidence under controlled content changes, without making a causal claim about real user engagement.

To study this setting, we introduce \emph{Counterfactual Behavior-Grounded Evidence Retrieval} (CBGER) and the CBGER-10K benchmark. CBGER-10K contains 5,000 factual--counterfactual pairs for 3,026 users, combining behavior histories from MicroLens with candidate video segments from FineVideo~\cite{ni2023microlens,farre2024finevideo}. Each pair changes only one focal segment while preserving the user, temporal position, and surrounding distractors. User and source identities are disjoint across splits, and the representations used to construct the benchmark are separated from the frozen CLIP features used by downstream models~\cite{radford2021learning}, reducing direct constructor--solver coupling.

We further introduce a compact behavior-conditioned model that decouples segment localization from video-level evidence existence. A temporal encoder produces user-conditioned segment scores for \textbf{Where}, while a parameter-free aggregation of these scores provides the \textbf{Whether} decision. Structured counterfactual learning combines factual localization, pairwise factual--counterfactual ordering, local intervention supervision, and invariance on unchanged distractors.

Experiments with adapted personalized-highlight and temporal-grounding baselines reveal a gap hidden by standard localization evaluation. CBGER achieves $0.4432$ MRR, $0.6977$ Pair Accuracy, and $0.6987$ Intervention Consistency. Compared with QD-DETR~\cite{moon2023query}, the MRR difference is not statistically significant, yet CBGER improves Pair Accuracy by $0.1103$ (95\% CI $[0.0765,0.1430]$). Thus, models with similar temporal localization quality can differ substantially in their ability to verify personalized evidence and respond to its removal.

Our contributions are:
\begin{itemize}[leftmargin=*]
\item We formulate personalized video evidence retrieval as separate \textbf{Where} and \textbf{Whether} decisions together with counterfactual intervention consistency.
\item We introduce CBGER-10K, a controlled benchmark of 5,000 factual--counterfactual pairs for 3,026 users with single-slot interventions and user- and source-disjoint splits.
\item We propose a compact behavior-conditioned framework with structured counterfactual learning that jointly trains localization, evidence ordering, local intervention response, and distractor invariance.
\item Across five adapted personalized-highlight and temporal-grounding baselines, we show that localization quality alone does not characterize personalized evidence reliability. Our CBGER-10K dataset and CBGER code are available at \url{https://anonymous.4open.science/r/cbger-111E}.
\end{itemize}

%% file: sec/related_work.tex
\section{Related Work}
\label{sec:related}

\textbf{Temporal grounding and personalized highlights.}
Temporal video grounding localizes language-relevant moments in untrimmed videos, progressing from proposal-based matching~\cite{gao2017tall,hendricks2017localizing,krishna2017dense} to transformer-based retrieval and unified grounding--highlight formulations~\cite{lei2021detecting,moon2023query,lin2023univtg}. Recent methods further improve temporal interaction, boundary modeling, foreground discrimination, and efficient localization~\cite{jang2023eatr,xiao2024uvcom,sun2024trdetr,cao2025flashvtg,sun2025mqvtg}. Video highlight detection similarly ranks salient moments, while personalized variants condition temporal relevance on viewer preferences or histories~\cite{gygli2016video2gif,badamdorj2022contrastive,chen2021personalized,yeh2023meta,lin2023collaborative}. These approaches are closely related to our \emph{Where} dimension, but primarily optimize which moment is most relevant within a video. CBGER additionally evaluates whether the localized content provides sufficient behavior-supported evidence at the video level.

\textbf{Negative-aware grounding, explainable recommendation, and counterfactual evaluation.}
Recent grounding work has begun to model missing correspondences explicitly. SHINE introduces semantically plausible hard negative queries~\cite{cheng2024shine}, while Negative-Aware Video Moment Retrieval requires models to reject queries whose events are absent from the video~\cite{flanagan2025moment}. These studies are related to our \emph{Whether} dimension, but operate on query--video correspondence rather than user-conditioned recommendation evidence. In parallel, explainable recommendation seeks human-interpretable reasons for item suggestions~\cite{zhang2020explainable}, and REASONER provides multimodal explanation labels for video recommendation~\cite{chen2023reasoner}. Counterfactual and grounded video evaluation further tests whether predictions depend on relevant visual or temporal evidence rather than shortcuts~\cite{zhang2024tv,bansal2024videocon,howard2024socialcf,xiao2024groundedqa}. Our setting combines these directions: a behavior-derived user profile defines personalized evidence, and matched factual--counterfactual videos replace one focal segment while holding the surrounding timeline fixed. This enables joint evaluation of localization, evidence existence, and local intervention consistency.

\textbf{Behavior-conditioned multimodal benchmarks.}
MicroLens provides large-scale user--micro-video interactions and multimodal content for recommendation research~\cite{ni2023microlens}, while FineVideo provides diverse temporally structured video content~\cite{farre2024finevideo}. Foundation models also enable scalable semantic annotation and temporal structure discovery~\cite{dvornik2023stepformer}. CBGER-10K builds on these resources but targets a different diagnostic question: whether temporally localized video content is supported by a user's observed behavior. Its automatically constructed labels are treated as silver supervision, and matched factual--counterfactual pairs together with user- and source-disjoint splits are used to evaluate personalized evidence reasoning under controlled content changes.

%% file: sec/task_dataset.tex
\section{Task and the CBGER-10K Benchmark}
\label{sec:data}

\subsection{Task Definition}

For a user $u$, let $H_u=(j_1,\ldots,j_M)$ denote the chronological consumption history and let $p_u$ be a behavior profile derived from that history. A candidate video is represented as a timeline of $N$ segments,
\begin{equation}
V=\{(x_i,[a_i,b_i])\}_{i=1}^{N}.
\end{equation}
Given $(p_u,V)$, a model predicts segment-level evidence scores
$S(u,V)=\{s_i\}_{i=1}^{N}$ and a video-level evidence-existence score $q(u,V)$.

\begin{figure*}[t]
\centering
\includegraphics[width=\textwidth]{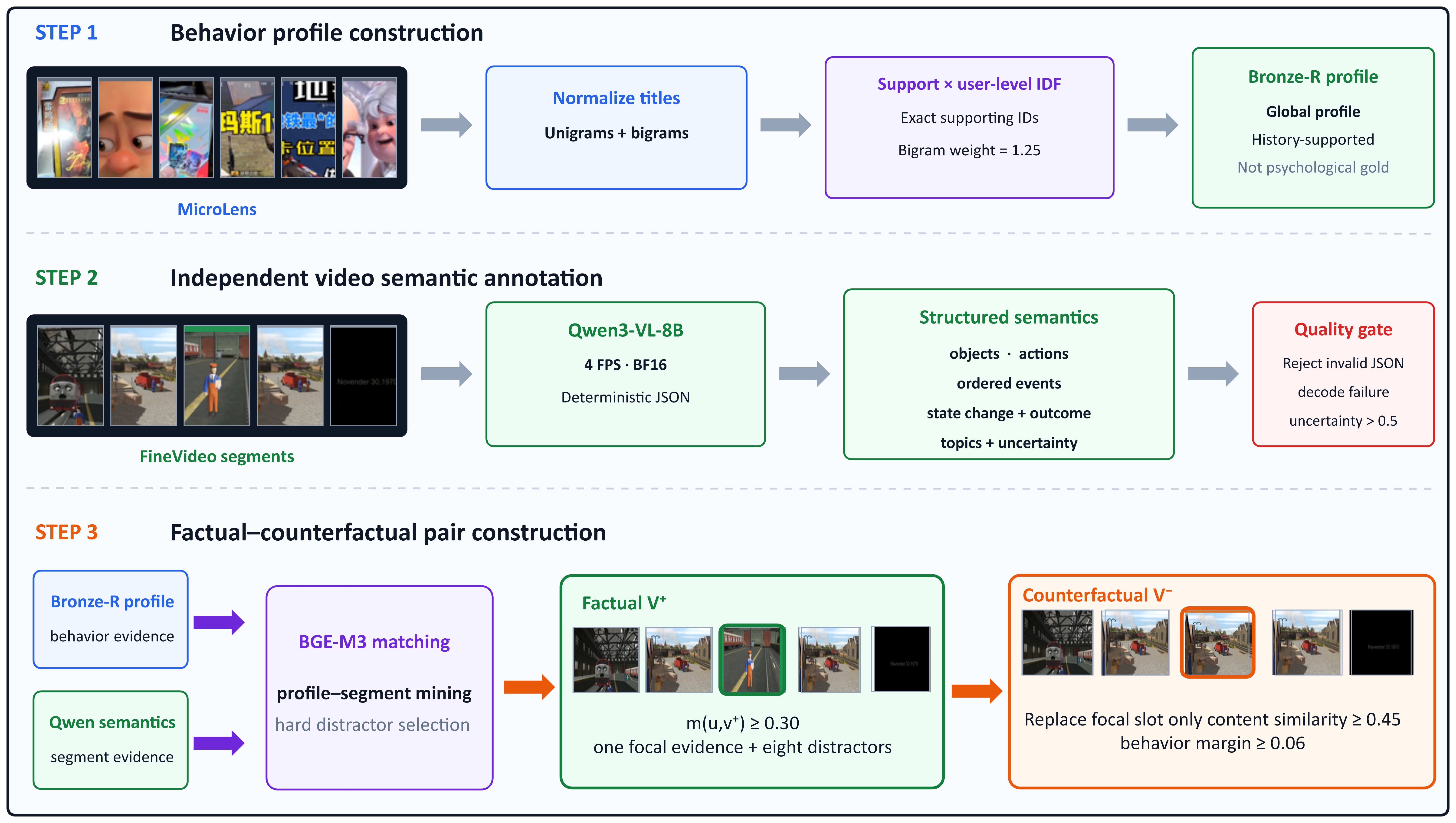}
\caption{\textbf{CBGER-10K construction pipeline.} MicroLens histories are converted into behavior-supported profiles. Eligible FineVideo intervals are independently annotated with structured visual semantics and matched to profiles to identify factual evidence. Each factual timeline contains one focal evidence segment and eight hard distractors. Its counterfactual preserves the user, focal temporal slot, and distractors while replacing only the focal segment with a semantically related but behavior-weaker content neighbor.}
\label{fig:data-pipeline}
\end{figure*}

CBGER-10K evaluates two prediction capabilities and one counterfactual consistency criterion. For a factual timeline $V^+$ with designated evidence segment $y$, \textbf{Where} evaluates whether $s_y$ is ranked above the distractor segments. Given its matched counterfactual $V^-$, \textbf{Whether} requires
\begin{equation}
q(u,V^+) > q(u,V^-).
\label{eq:pair-order}
\end{equation}
Finally, \textbf{Intervention Consistency} tests whether replacing the factual evidence at the same temporal slot decreases its local evidence score,
\begin{equation}
s_y(u,V^+) > s_y(u,V^-).
\label{eq:intervention}
\end{equation}
The benchmark therefore distinguishes successful localization from recognizing whether personalized evidence is present and whether predictions depend on that evidence. Figure~\ref{fig:data-pipeline} illustrates the pipeline for constructing CBGER-10K.

\subsection{Behavior Profiles and Candidate Videos}

\paragraph{Behavior profiles.}
MicroLens provides timestamped user--item interactions and multimodal micro-video metadata~\cite{ni2023microlens}. We use chronological interactions and video titles to construct behavior-supported user profiles; unobserved items are not treated as negatives. After normalization, title unigrams and bigrams are linked to the exact history items that support them. Their user-level inverse document frequency is
\begin{equation}
\operatorname{idf}(a)=
\log\frac{|\mathcal U|+1}{\operatorname{df}_{\mathcal U}(a)+1},
\end{equation}
and phrase support is
\begin{equation}
c_u(a)=|H_u(a)|\operatorname{idf}(a)\rho(a),
\end{equation}
where $\rho(a)=1.25$ for bigrams and $1$ otherwise. The highest-support phrases and their associated history items form the global behavior profile $p_u$. These profiles are operational summaries of observed consumption rather than psychological ground-truth interests.

\paragraph{Candidate videos.}
FineVideo provides Creative-Commons videos with temporal intervals, transcripts, and structured descriptions~\cite{farre2024finevideo}. We use its media and temporal provenance but do not use FineVideo category overlap as evidence supervision. Eligible intervals have durations between 1 and 12 seconds. FineVideo source videos are assigned to train, validation, and test splits before annotation or matching.

\subsection{Semantic Annotation and Matching}

Each eligible FineVideo interval is annotated by Qwen3-VL-8B-Instruct~\cite{qwen3vl2025} using a fixed 4-FPS BF16 configuration selected on a held-out configuration study. The structured annotation contains objects, actions, ordered events, state changes, outcomes, topics, and an uncertainty estimate. Invalid outputs and annotations with uncertainty above $0.5$ are discarded, yielding 7,651 valid intervals. We serialize the retained semantics as
\begin{align}
\phi(v)=&\;\texttt{Topics}(T_v);\texttt{Objects}(O_v);
\texttt{Actions}(A_v);\nonumber\\
&\texttt{Events}(E_v);\texttt{Outcome}(Y_v).
\end{align}

User profiles and interval semantics are independently encoded with BGE-M3~\cite{chen2024bgem3}:
\begin{equation}
h_u=\frac{f(\psi(u))}{\|f(\psi(u))\|_2},
\qquad
z_v=\frac{f(\phi(v))}{\|f(\phi(v))\|_2},
\end{equation}
with construction similarity
\begin{equation}
m(u,v)=h_u^\top z_v.
\end{equation}
A segment may serve as factual evidence only if $m(u,v)\ge0.30$.

Qwen3-VL and BGE-M3 are used only for benchmark construction. Downstream models receive frozen CLIP features~\cite{radford2021learning}; construction embeddings and matching scores are never exposed to the solver. This representation separation reduces direct constructor--solver leakage, although it does not remove potential teacher bias.

\subsection{Factual--Counterfactual Pair Construction}

Construction is performed independently within each split. For each user, factual candidates are ranked by $m(u,v)$ subject to diversity constraints: the same user--segment pair is not repeated, each evidence segment is reused at most three times, and each user contributes at most two pairs.

For a selected factual segment $v^+$, we mine a semantically related counterfactual replacement from its nearest content neighbors. Content similarity is
\begin{equation}
c(v^+,v)=z_{v^+}^{\top}z_v.
\end{equation}
A replacement $v^-$ must satisfy
\begin{equation}
c(v^+,v^-)\ge0.45,
\qquad
m(u,v^+)-m(u,v^-)\ge0.06.
\label{eq:cf-constraints}
\end{equation}
Thus, $v^-$ remains close in content while providing weaker support for the same user profile. Distractor segments are additionally required to satisfy
\begin{equation}
m(u,d)\le m(u,v^+)-0.03.
\end{equation}

Each factual timeline contains nine unique segments: one focal evidence segment and eight hard distractors. The focal position is randomly sampled with a fixed seed. The matched counterfactual preserves the user profile, history, temporal slot, and all eight distractors, and replaces only $v^+$ with $v^-$. This single-slot intervention minimizes changes outside the designated evidence while allowing direct comparison of factual and counterfactual predictions.

All records retain source identifiers, temporal intervals, construction scores, counterfactual links, random seeds, and integrity hashes for reproducibility. We emphasize that these are automatically constructed silver labels: the intervention measures model dependence on designated personalized evidence, not the causal effect of video content on observed user engagement.

\subsection{Splits and Benchmark Statistics}

Table~\ref{tab:data} summarizes the key statistics of the CBGER-10K benchmark. Users are assigned to train, validation, and test partitions using a stable 70/10/20 split. FineVideo source videos are independently partitioned before candidate selection, resulting in zero user or source overlap across splits. Structural validation checks factual--counterfactual links, shared histories and distractors, single-slot replacement, temporal intervals, matching constraints, and feature coverage; all 5,000 pairs pass these checks.

\begin{table}[t]
\centering
\caption{CBGER-10K benchmark statistics. Each pair contributes one factual and one counterfactual record.}
\label{tab:data}
\small
\resizebox{\columnwidth}{!}{%
\begin{tabular}{lrrrr}
\toprule
Statistic & Train & Val. & Test & Total\\
\midrule
Records & 7,000 & 1,000 & 2,000 & 10,000\\
Pairs & 3,500 & 500 & 1,000 & 5,000\\
Unique users & 2,103 & 307 & 616 & 3,026\\
Factual records & 3,500 & 500 & 1,000 & 5,000\\
Counterfactual records & 3,500 & 500 & 1,000 & 5,000\\
\bottomrule
\end{tabular}}
\end{table}

\subsection{Evaluation Metrics}

We report the three benchmark axes separately. \textbf{Where} is evaluated on factual records using MRR and NDCG@$k$, treating the designated factual segment as the relevant item.

\textbf{Whether} is measured by Pair Accuracy,
\begin{equation}
A_{\rm pair}
=
\frac{1}{|\mathcal P|}
\sum_{(V^+,V^-)\in\mathcal P}
\mathbf 1
\left[
q(u,V^+)>q(u,V^-)
\right].
\end{equation}

\textbf{Intervention Consistency} measures whether the score at the fixed focal slot decreases after replacement:
\begin{equation}
A_{\rm int}
=
\frac{1}{|\mathcal P|}
\sum_{(V^+,V^-)\in\mathcal P}
\mathbf 1
\left[
s_y(u,V^+)>s_y(u,V^-)
\right].
\end{equation}
Reporting these metrics separately is important because a model may localize the factual segment accurately while still failing to distinguish evidence-present from matched evidence-weakened videos.

%% file: sec/method.tex
\section{Method}
\label{sec:method}

\subsection{Overview}

\begin{figure*}[t]
\centering
\includegraphics[width=\textwidth]{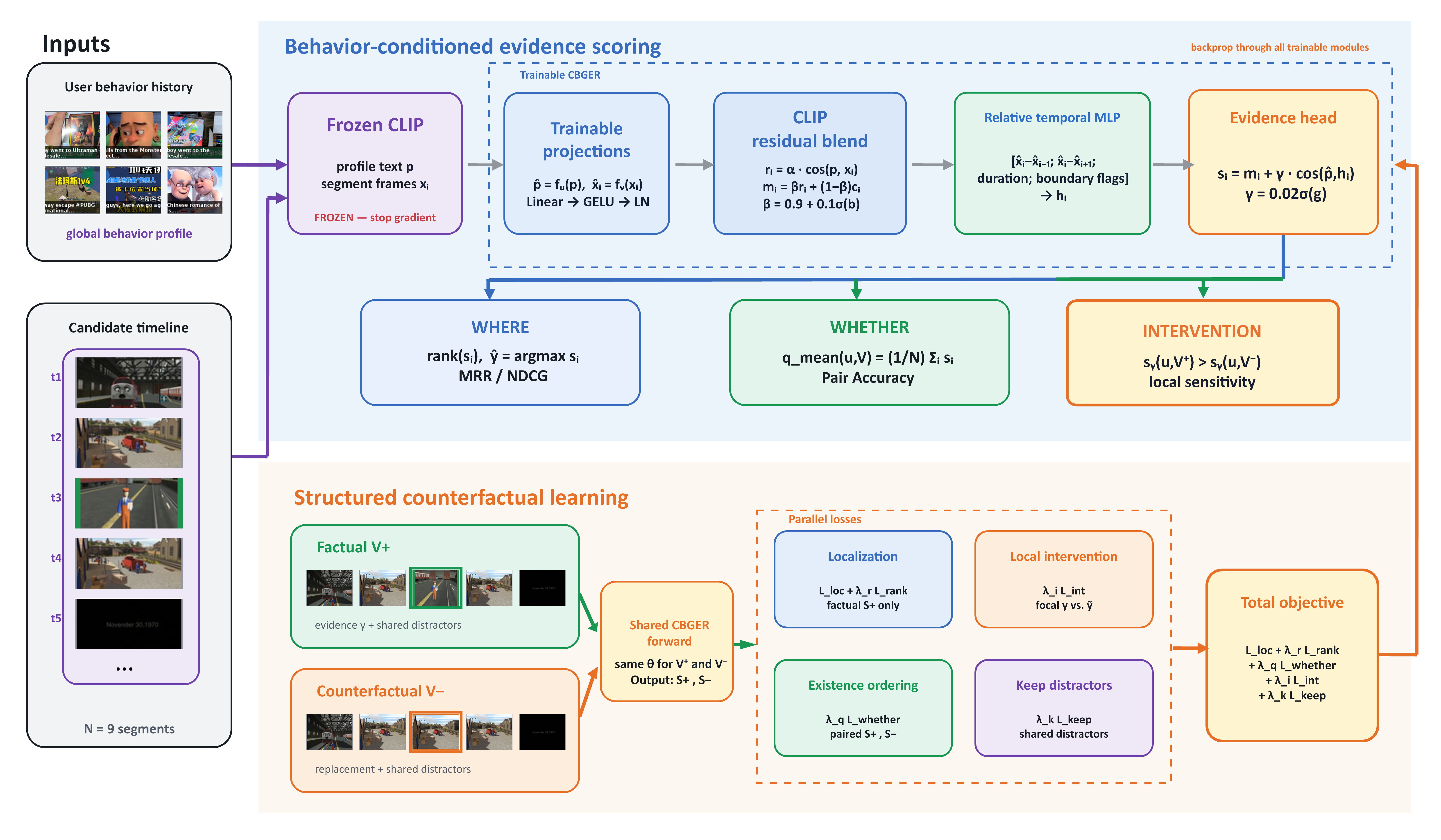}
\caption{\textbf{Behavior-conditioned evidence scoring and structured counterfactual learning.} Frozen CLIP features are processed by lightweight projections, a CLIP residual branch, and a relative-temporal MLP to produce segment evidence logits. Their ranking answers \textbf{Where}, while parameter-free mean pooling answers \textbf{Whether}. Factual $V^+$ and matched counterfactual $V^-$ share parameters and are jointly supervised by localization, existence-ordering, local-intervention, and distractor-invariance losses. Gradients update only the lightweight CBGER scorer; CLIP remains frozen.}
\label{fig:model}
\end{figure*}

We propose a compact behavior-conditioned evidence model for CBGER as shown in Figure~\ref{fig:model}. Given a user behavior profile and a nine-segment candidate timeline, the model produces segment-level evidence scores for \textbf{Where} and a video-level evidence score for \textbf{Whether}. Frozen CLIP encoders provide profile and segment representations~\cite{radford2021learning}. A lightweight trainable scorer combines a CLIP similarity residual with a small relative-temporal correction. During training, the same model is applied to matched factual and counterfactual timelines. Structured objectives supervise factual localization, video-level factual--counterfactual ordering, the response at the edited slot, and invariance of unchanged distractors.

\subsection{Behavior-Conditioned Evidence Scoring}

Let $x_i\in\mathbb R^{d_c}$ denote the frozen CLIP representation of segment $i$, and let $p\in\mathbb R^{d_c}$ denote the frozen CLIP text representation of the serialized behavior profile. We first compute the raw CLIP similarity
\begin{equation}
r_i=\alpha\,\operatorname{cos}(p,x_i),
\label{eq:raw-clip}
\end{equation}
where $\alpha$ is a learnable scale. Lightweight projections map both modalities to a shared $d$-dimensional space:
\begin{equation}
\begin{aligned}
\hat{p}
&= \operatorname{LN}\!\left(
   \operatorname{GELU}(W_p p + b_p)
   \right), \\
\hat{x}_i
&= \operatorname{LN}\!\left(
   \operatorname{GELU}(W_x x_i + b_x)
   \right).
\end{aligned}
\end{equation}
The projected similarity $c_i=\alpha\,\operatorname{cos}(\hat p,\hat x_i)$ is combined with the raw CLIP score through a constrained residual weight:
\begin{equation}
m_i=\beta r_i+(1-\beta)c_i,
\qquad
\beta=0.9+0.1\sigma(b).
\label{eq:clip-residual}
\end{equation}
This parameterization preserves the pretrained CLIP geometry while allowing a small task-specific correction.

To encode local temporal structure without self-attention, we construct
\begin{equation}
\begin{aligned}
d_i = \big[
&\hat{x}_i-\hat{x}_{i-1};
 \hat{x}_i-\hat{x}_{i+1}; 
&\ell_i;
 b_i^{\mathrm{L}};
 b_i^{\mathrm{R}}
\big], \\
h_i &= \operatorname{MLP}_{\mathrm{temp}}(d_i).
\end{aligned}
\label{eq:relative-temporal}
\end{equation}
where $\ell_i$ is the normalized segment duration and $b_i^{\rm L},b_i^{\rm R}$ indicate timeline boundaries. The final evidence logit is
\begin{equation}
s_i=m_i+\gamma\,\operatorname{cos}(\hat p,h_i),
\qquad
\gamma=0.02\sigma(g).
\label{eq:segment-score}
\end{equation}
The bounded gate makes temporal context a light correction rather than a replacement for semantic similarity. The localized prediction is
\begin{equation}
\hat y=\arg\max_i s_i,
\end{equation}
and ranking $\{s_i\}_{i=1}^{N}$ defines the \textbf{Where} pathway.

\subsection{Video-Level Evidence Estimation}

Localization does not by itself establish that valid evidence exists: after evidence replacement, a maximum operator must still select the strongest distractor. We therefore use parameter-free mean pooling for the \textbf{Whether} score:
\begin{equation}
q(u,V)=\frac{1}{N}\sum_{i=1}^{N}s_i.
\label{eq:mean-pool}
\end{equation}
This rule evaluates support over the complete controlled timeline, introduces no additional parameters, and is compared with alternative aggregators in the ablation study.

\subsection{Intervention Consistency}

The intervention task measures whether the score of the edited focal moment decreases when its supporting evidence is replaced. For a factual--counterfactual pair $(V^+,V^-)$ with the intervention applied at slot $y$, we define
\begin{equation}
I(u,V^+,V^-)
=\mathbf{1}\!\left[s_y(u,V^+)>s_y(u,V^-)\right].
\label{eq:intervention-correct}
\end{equation}
The dataset-level Intervention score is the mean of this indicator over all $M$ matched pairs:
\begin{equation}
\operatorname{Intervention}
=\frac{1}{M}\sum_{n=1}^{M}
I(u_n,V_n^+,V_n^-).
\label{eq:intervention-metric}
\end{equation}
Unlike \textbf{Whether}, which compares the video-level scores $q(u,V^+)$ and $q(u,V^-)$, Intervention evaluates the model's local response at the known edited slot.

\subsection{Structured Counterfactual Learning}

Let $V^+$ be a factual timeline with evidence at slot $y$, and let $V^-$ be its matched counterfactual obtained by replacing only that slot. Both timelines are processed with shared parameters, producing $S^+=\{s_i^+\}$ and $S^-=\{s_i^-\}$.

\paragraph{Factual localization.}
We supervise the evidence position using temperature-scaled cross entropy and a hard-negative ranking margin:
\begin{equation}
\mathcal L_{\rm loc}
=-log\frac{\exp(s_y^+/\tau)}{\sum_{i=1}^{N}\exp(s_i^+/\tau)},
\end{equation}
\begin{equation}
\mathcal L_{\rm rank}
=\frac{1}{N-1}\sum_{i\neq y}[\delta_{\mathrm{rank}}-s_y^++s_i^+]_+.
\end{equation}
where $\delta_{\mathrm{rank}} > 0$ is the ranking margin.
\paragraph{Video-level evidence ordering.}
The factual timeline should provide stronger support than its counterfactual:
\begin{equation}
\mathcal L_{\rm whether}
=\log\!\left(1+\exp\!\left[-\frac{q(u,V^+)-q(u,V^-)}{\tau_q}\right]\right).
\label{eq:whether-loss}
\end{equation}
This objective directly matches the factual--counterfactual ordering measured by Pair Accuracy.

\paragraph{Local intervention and distractor invariance.}
Because only the focal slot is edited, its score should decrease while the unchanged distractors should remain stable:
\begin{equation}
\mathcal L_{\rm int}=[\delta_{\mathrm{ink}}-s_y^+ + s_y^-]_+,
\label{eq:int-loss}
\end{equation}
\begin{equation}
\mathcal L_{\rm keep}
=\frac{1}{|D|}\sum_{j\in D}(s_j^+-s_j^-)^2,
\end{equation}
where $\delta_{\mathrm{ink}} > 0$ is the intervention margin and $D$ indexes the eight shared distractor slots. The latter term prevents a trivial global shift of all counterfactual scores.

The complete objective is
\begin{equation}
\mathcal L
=\mathcal L_{\rm loc}
+\lambda_r\mathcal L_{\rm rank}
+\lambda_q\mathcal L_{\rm whether}
+\lambda_i\mathcal L_{\rm int}
+\lambda_k\mathcal L_{\rm keep}.
\label{eq:total-loss}
\end{equation}
All losses are computed from the shared factual and counterfactual forward passes. Gradients update the projections, residual and temporal gates, relative-temporal MLP, while the CLIP image and text encoders remain frozen.

%% file: sec/experiments.tex
\section{Experiments}
\label{sec:experiments}

\subsection{Experimental Setup}

All methods use the same frozen CBGER-10K train/validation/test splits and CLIP ViT-B/32 visual and text representations~\cite{radford2021learning}. CBGER uses hidden dimension $d=256$, a two-layer relative-temporal MLP, AdamW with learning rate $2\times10^{-4}$, and weight decay $10^{-4}$. Hyperparameters and checkpoints are selected on the validation set; the test set is used only for final evaluation. Results are averaged over three random seeds and reported as mean$\pm$standard deviation.

We compare with six baselines: frozen CLIP profile--segment similarity; PR-Net for personalized highlight detection~\cite{chen2021personalized}; QD-DETR~\cite{moon2023query}; TR-DETR~\cite{sun2024trdetr}; FlashVTG~\cite{cao2025flashvtg}; and MQVTG~\cite{sun2025mqvtg}. For temporal-grounding baselines, the user's behavior profile replaces the original language query. All methods use the same candidate timelines, downstream representations, and data splits, and are evaluated using a single implementation of the CBGER metrics.

\subsection{Main Results}

\begin{table*}[t]
\centering
\caption{Results on CBGER-10K. PairAcc measures factual--counterfactual evidence ordering, and Intervention measures the local score response at the replaced evidence slot. Best results are in bold and second-best are underlined.}
\label{tab:main}
\small
\resizebox{\textwidth}{!}{%
\begin{tabular}{lcccccc}
\toprule
Method & MRR & NDCG@1 & NDCG@3 & NDCG@5 & PairAcc & Intervention\\
\midrule
Frozen CLIP similarity & .4127$\pm$.0000 & .2060$\pm$.0000 & .3588$\pm$.0000 & .4461$\pm$.0000 & .6320$\pm$.0000 & .6320$\pm$.0000\\
PR-Net & .3850$\pm$.0047 & .1627$\pm$.0098 & .3368$\pm$.0130 & .4271$\pm$.0151 & .5810$\pm$.0436 & .6233$\pm$.0150\\
QD-DETR & \underline{.4231$\pm$.0011} & \underline{.2063$\pm$.0051} & \underline{.3784$\pm$.0018} & \underline{.4662$\pm$.0016} & .5873$\pm$.0080 & \underline{.6517$\pm$.0210}\\
TR-DETR & .4125$\pm$.0236 & .2043$\pm$.0154 & .3630$\pm$.0357 & .4501$\pm$.0380 & \underline{.6320$\pm$.0212} & .6527$\pm$.0202\\
FlashVTG & .4154$\pm$.0061 & .2013$\pm$.0099 & .3736$\pm$.0081 & .4549$\pm$.0060 & .6207$\pm$.0153 & .6327$\pm$.0029\\
MQVTG & .4112$\pm$.0137 & .1940$\pm$.0148 & .3624$\pm$.0146 & .4552$\pm$.0223 & .6073$\pm$.0257 & .6380$\pm$.0118\\
\midrule
\textbf{CBGER} & \textbf{.4432$\pm$.0122} & \textbf{.2363$\pm$.0099} & \textbf{.4008$\pm$.0187} & \textbf{.4901$\pm$.0133} & \textbf{.6977$\pm$.0070} & \textbf{.6987$\pm$.0035}\\
\bottomrule
\end{tabular}}
\end{table*}

As shown in Table~\ref{tab:main}, CBGER achieves the best performance on all three evaluation axes. The most informative comparison is with QD-DETR. Their localization performance is close (MRR $.4432$ vs. $.4231$), but CBGER improves PairAcc by $0.1103$ and Intervention Consistency by $0.0470$. TR-DETR obtains the strongest baseline PairAcc at $.6320$, still $6.57$ points below CBGER.

We assess significance using 10,000 paired bootstrap replicates clustered by user, preserving all predictions associated with each sampled user. Relative to QD-DETR, the MRR difference is not significant ($+0.0201$, 95\% CI $[-0.0027,0.0433]$, $p=.0832$), whereas the PairAcc gain is substantial ($+0.1103$, 95\% CI $[0.0765,0.1430]$, $p<10^{-4}$) and the Intervention gain is also significant ($+0.0470$, 95\% CI $[0.0166,0.0772]$, $p=.0022$). Thus, two models can be statistically similar in temporal localization while differing markedly in personalized evidence existence and intervention response.

\subsection{Where--Whether Ablation}

We next separate the contribution of structured counterfactual learning from the video-level aggregation rule.

\begin{table}[t]
\centering
\caption{Structured counterfactual learning (CF) and mean evidence pooling.}
\label{tab:factorial}
\small
\resizebox{\columnwidth}{!}{%
\begin{tabular}{ccccc}
\toprule
CF & Mean & MRR & PairAcc & Intervention\\
\midrule
\xmark & \xmark & .4274$\pm$.0100 & .6230$\pm$.0046 & .6830$\pm$.0079\\
\xmark & \cmark & .4274$\pm$.0100 & .6810$\pm$.0085 & .6830$\pm$.0079\\
\cmark & \xmark & .4432$\pm$.0122 & .6297$\pm$.0085 & .6987$\pm$.0035\\
\cmark & \cmark & \textbf{.4432$\pm$.0122} & \textbf{.6977$\pm$.0070} & \textbf{.6987$\pm$.0035}\\
\bottomrule
\end{tabular}}
\end{table}

As shown in Table~\ref{tab:factorial}, structured counterfactual training increases MRR from $.4274$ to $.4432$ and Intervention from $.6830$ to $.6987$, showing that paired supervision improves both factual localization and sensitivity to the edited slot. Mean pooling does not alter segment rankings and therefore leaves MRR and Intervention unchanged, but improves PairAcc by $5.8$ points without counterfactual training and $6.8$ points with it. Under mean pooling, counterfactual training further improves PairAcc by $1.67$ points. The interaction between the two components is not statistically significant, indicating complementary rather than super-additive effects.

\subsection{How Should Evidence Existence Be Aggregated?}

To isolate the \textbf{Whether} decision, we keep CBGER segment scores fixed and vary only the video-level aggregation rule.

\begin{table}[t]
\centering
\caption{Video-level evidence aggregation with fixed segment scores.}
\label{tab:whether}
\small
\begin{tabular}{lc}
\toprule
Aggregation rule & PairAcc\\
\midrule
Max & .4860$\pm$.0200\\
Top-2 mean & .5660$\pm$.0166\\
Top-3 mean & .6160$\pm$.0177\\
Learned existence MLP & .4560$\pm$.0035\\
\textbf{Mean} & \textbf{.6977$\pm$.0070}\\
\bottomrule
\end{tabular}
\end{table}

As shown in Table~\ref{tab:whether}, max pooling performs poorly because each factual--counterfactual pair shares eight hard distractors; the maximum is therefore often determined by an unchanged segment and becomes insensitive to the focal replacement. Increasing the number of aggregated segments progressively improves PairAcc. A learned pointwise existence head also underperforms, indicating that additional capacity alone does not resolve the existence decision. Mean pooling provides the strongest performance while introducing no additional parameters.

\subsection{Qualitative Intervention Analysis}

Two illustrative intervention-success cases are provided in the Appendix. The baselines frequently identify visually plausible or topically related factual segments, yet fail to lower either the video-level score or the focal segment score after the supporting moment is replaced. CBGER instead preserves the factual--counterfactual ordering and produces a localized score decrease at the edited slot. These examples illustrate the quantitative finding that plausible temporal localization alone does not guarantee reliable evidence verification.

%% file: sec/limitations.tex
\section{Limitations and responsible interpretation}
CBGER-10K is a controlled diagnostic benchmark with several limitations. Its automatically constructed Qwen3-VL/BGE-M3 labels may contain semantic and matching biases; counterfactual replacements are behavior-weaker under the construction criteria, but are not guaranteed to be universally irrelevant. Because user histories come from MicroLens and candidate videos from FineVideo, the benchmark measures behavior-supported evidence rather than actual user exposure, preference, or causal engagement. The fixed nine-segment timelines and frozen CLIP backbone improve experimental control but limit generalization to unrestricted videos and other representations. Finally, PairAcc measures relative evidence ordering within matched pairs, not calibrated engagement probability; CBGER-10K should therefore be viewed as a diagnostic benchmark for personalized evidence reasoning rather than a simulation of deployed recommendation systems.

%% file: sec/conclusion.tex
\section{Conclusion}
We introduced counterfactual behavior-grounded evidence retrieval, which separates \textbf{Where} personalized video evidence occurs from \textbf{Whether} such evidence exists and whether model predictions respond consistently to its replacement. CBGER-10K provides 5,000 controlled factual--counterfactual pairs for evaluating these capabilities, and CBGER jointly learns localization, evidence existence, and intervention consistency through structured counterfactual supervision. Experiments show that strong temporal localization does not necessarily imply reliable personalized evidence existence, while CBGER consistently improves Pair Accuracy and Intervention Consistency over existing grounding and personalized-highlight baselines. More broadly, our results suggest that evaluating only where a model attends can hide failures in whether the retrieved content is supported by the benchmark’s behavior-derived evidence criteria; controlled interventions provide a practical way to expose this gap.